\documentclass{nature-modified}
\usepackage{xcolor}

\title{Haptic communication optimises joint decisions and \mbox{affords implicit confidence sharing}}

\author{Giovanni Pezzulo$^{1,*}$, Lucas Roche$^2$ \& Ludovic Saint-Bauzel$^2$}

\begin{document}

\maketitle

\begin{affiliations}
 \item Institute of Cognitive Sciences and Technologies, National Research Council, Via S. Martino della Battaglia 44, 00185 Rome, Italy
 \item Institut des Systemes Intelligents et de Robotique, Universit\'{e} Pierre et Marie Curie, 75005 Paris, France
 \item[*] Corresponding author (giovanni.pezzulo@istc.cnr.it)
\end{affiliations}

\begin{abstract}
Group decisions can outperform the choices of the best individual group members. Previous research suggested that optimal group decisions require individuals to communicate \emph{explicitly} (e.g., verbally) their confidence levels. Our study addresses the untested hypothesis that  \emph{implicit} communication using a sensorimotor channel -- haptic coupling -- may afford optimal group decisions, too. We report that haptically coupled dyads solve a perceptual discrimination task more accurately than their best individual members; and five times faster than dyads using explicit communication. Furthermore, our computational analyses indicate that the haptic channel affords implicit confidence sharing. We found that dyads take leadership over the choice and communicate their confidence in it by modulating both the timing and the force of their movements. Our findings may pave the way to negotiation technologies using fast sensorimotor communication to solve problems in groups.
\end{abstract}

\newpage

\textcolor{black}{We often make important decisions in groups, such as when we decide a travel destination with a group of friends or peer-review papers.} Group decisions can sometimes outperform the choices of the best individual group members; for example, during logical problems \cite{moshman1998}, numerical \cite{bang2017} or perceptual tasks \cite{bahrami2010}. 

There is strong consensus that effective group decisions require group members to share their degree of confidence in their individual choices \cite{bahrami2010,Bahrami2012,Fusaroli2012,Haller2018,koriat2012,Sorkin2001}. This allows weighting individual choices according to their relative confidence levels, following principles of optimal (Bayesian) multisensory integration \cite{Ernst2002,Kording2006,Doya2007}. It has been assumed so far that \emph{explicit} communicative channels, such as verbal communication \cite{bahrami2010} or visual confidence reports \cite{Bahrami2012}, are required to share confidence levels -- and more broadly, negotiate optimal decisions \cite{Navajas2018}. \textcolor{black}{This idea is in keeping with a long tradition in communication theory and cognitive psychology that emphasises the importance of communicating intentions and metacognitive confidence levels explicitly (e.g., verbally) \cite{Metcalfe1996,Sperber1995}.}


\textcolor{black}{However, during ecologically realistic interactions, dyads also use \emph{implicit}, sensorimotor communication channels to improve coordination and achieve joint goals \cite{Pezzulo2018c,Sebanz2006}. For example, during joint grasping or joint pressing tasks, dyads modulate (e.g., amplify) the kinematics of their finger and arm movements to make the trajectory of their movements less variable and hence more predictable \cite{Vesper2011a}, to make their intentions (e.g., what object they intend to grasp and when) easier to infer by coactors \cite{Candidi2015,Curioni2017,Pezzulo2013,Sacheli2013,Sartori2009,Vesper2014}, or to dynamically negotiate leader-follower roles \cite{Konvalinka2010,Noy2011,Skewes2015}. Moreover, in the same tasks, participants tend to implicitly imitate each other's actions \cite{Era2020,Gandolfo2019} -- a mechanism suggested to promote group affiliation \cite{SalazarKampf2018}. During joint pulling tasks, haptically coupled dyads amplify their force to improve their coordination \cite{Wel2010}. As a result, sensorimotor communication during jointly executed tasks can improve performance and the quality of execution \cite{D'Ausilio2012,Ganesh2014,Groten2010,Malysz2013,Masumoto2015,Pezzulo2017,Reed2008,Takagi2018,Takagi2019}. This suggests the untested hypothesis that \emph{implicit} communication based on sensorimotor (e.g., haptic) channels  -- which is faster and cheaper in cognitive load than explicit communication --  may be sufficient to optimize group decisions and share confidence levels.} 

To test this hypothesis, we adapted a previous task designed to study optimal group decisions using verbal communication \cite{bahrami2010} -- but we allowed participants to communicate only via a sensorimotor (haptic) channel. In our study, dyads (couples of individuals) make a series of \emph{individual} decisions and then -- if they disagree -- \emph{group} (consensus) decisions, about which of two sequentially presented stimuli contains an oddball target. During the group decisions, the dyads control coupled haptic devices with one degree-of-freedom, to jointly move the end effectors towards one of two (left or right) extreme positions, corresponding to their two choices. 

We report three \textcolor{black}{three main findings}. First, we show that haptic communication allows dyads to optimize group decisions and outperform the accuracy of individual participants (having similar sensitivity levels), akin to consensus reached using verbal communication \cite{bahrami2010} -- but five times faster. Second, our computational analysis indicates that the haptic channel affords sharing confidence levels during the group choice -- albeit in an implicit form. Indeed, the same computational (Bayesian) scheme explains group choices using both explicit (verbal) and implicit (haptic) communication, in terms of (weighted) confidence sharing. \textcolor{black}{Third}, our analyses indicate that dyads take leadership over the choice and communicate their confidence in it by manipulating both the timing and force of their movements -- hence exploiting the haptic channel in full to optimise their joint performance.

\section*{Results}

\begin{figure}[h]
\centering
  \includegraphics[width=.9\linewidth]{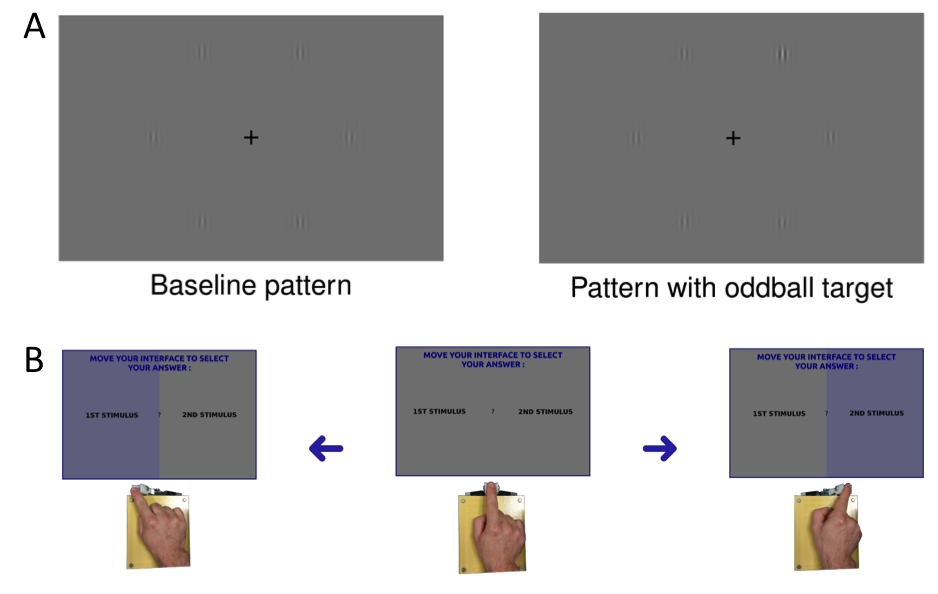}
  \caption{Experimental setup. (A) Example experimental stimuli, without (left) or with (right) an oddball target. (B) Graphical illustration of the (left-right) decisions using the haptic interface. Participants are presented with a choice between the first and second stimulus (center panel) and have to move the haptic interface to the left (left panel) or right (right panel). The setup is the same for both individual and group decisions; but while during individual decisions the haptic interfaces of the two participants are not connected, they are connected during group decisions. See the main text for details.}
  \label{fig:figure1}		
\end{figure}

Couples of participants (dyads) are presented with a series of two stimuli, one of which containing an oddball target. Each participant sees the (identical) stimuli on a different computer screen (Figure~\ref{fig:figure1}). During the first, \emph{individual} decision phase, each participant indicates which of the stimuli (first or second) contains the oddball target, by moving a cursor to the left or right response \textcolor{black}{area}. Each participant controls his or her cursor independently, using a haptic device. In this phase, the devices of the two participants are not coupled and participants cannot communicate. If the individual decisions are identical, the trial ends. If they differ, a second \emph{consensus} decision phase begins, in which participants make the same decision as above, but jointly control the cursor trajectory (i.e., the cursor trajectory is an average of the individual trajectories). Different from the first phase, participants see the joint (not the individual) cursor trajectory on their screens. Furthermore, the participants' haptic devices are coupled and permit sensing the amount of force the co-actor applies to his or her device. 

%

\subsection{Haptic communication optimizes group decisions alike explicit communication -- but is much faster}

\begin{figure}[h]
\center
\includegraphics[width=.9\linewidth]{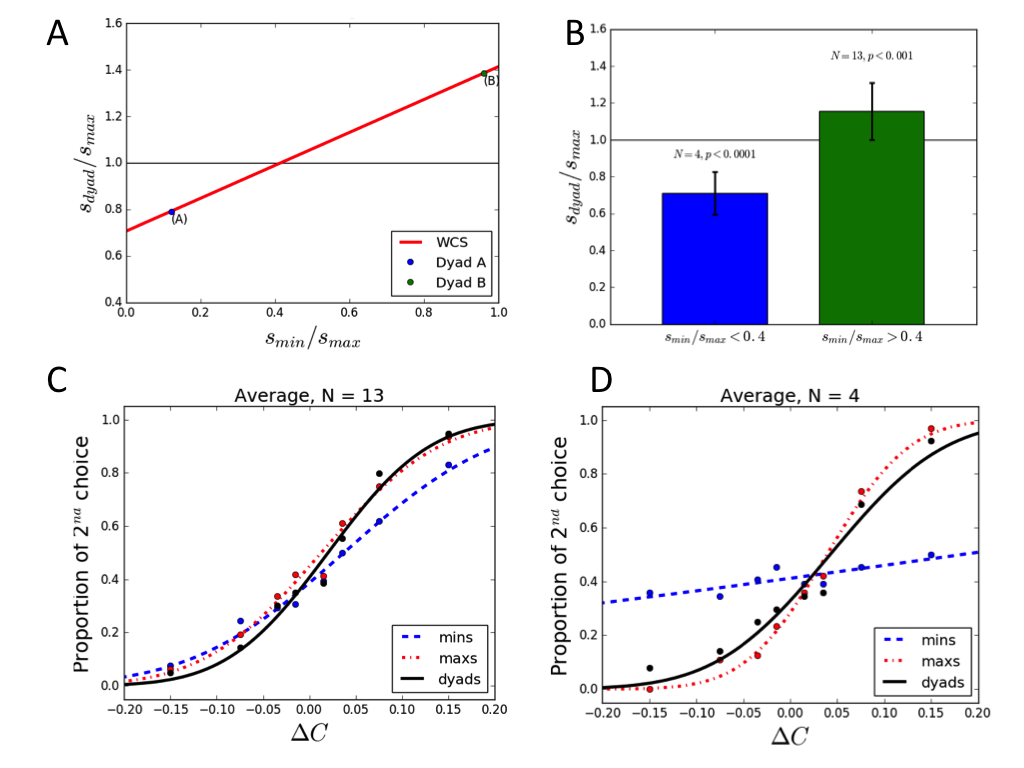}
\caption{Experimental results. (A) Theoretical prediction of the WCS model: dyads outperform their best individual members ($s_{dyad}/s_{max} > 1$) if members have similar sensitivities ($s_{min}/s_{max} > 0.4$). (B) Performance of dyads whose members have similar (green) or different (blue) sensitivities. (C,D) Average psychometric functions of the worst (blue) and best (red) individual members, compared to the dyad (black). Dots are the average percentage of 2$^{nd}$ stimuli chosen as answer, for each contrast difference in the experiment. Lines are fitted cumulative Gaussian functions. Steeper slopes correspond to higher sensitivities.}
\label{fig:figure2}
\end{figure}

We fitted the response data using the \emph{weighted confidence sharing (WCS)} model, which successfully explained group decisions using verbal communication \cite{bahrami2010}. The WCS model assumes that group decisions weight individual choices according to their relative confidence levels. It makes the theoretical prediction that when the ratio of sensitivities of dyad members is greater than 0.4 (i.e., participants have similar sensitivities), then the dyad will outperform each individual (Figure~\ref{fig:figure2}A). The opposite happens if the ratio is lower than 0.4 (i.e., participants have different sensitivities). This prediction was confirmed, with dyads whose members had similar sensitivities ($s_{min}/s_{max} > 0.4$) performing significantly better than their best members (\textit{t(13)=3.94, p$<$0.001}) and dyads whose members had different sensitivities ($s_{min}/s_{max} < 0.4$) performed significantly worse than their best members (\textit{t(4)=-9.89, p$<$0.0001}) (Figure~\ref{fig:figure2}B-D).

\begin{figure}[h]
\center
\includegraphics[width=.9\linewidth]{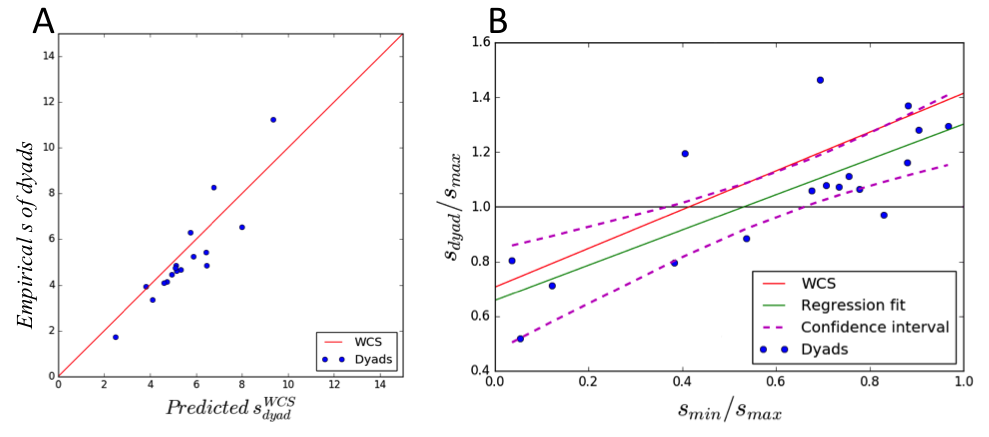}
\caption{Computational analyses. (A) Correlation between dyads' sensitivities observed in the experiment and those predicted by the WCS model. Blue dots: data points (mean data of each dyad over the 8 blocks); red line: theoretical prediction of the WCS model. (B) Correlation between dyads' sensitivity, relative to the best member of the dyad. Green and purple lines: linear regression model fitted on the data points and 95\% confidence interval, respectively.}
\label{fig:figure3}
\end{figure}

Furthermore, the slopes of the dyads' psychometric functions and those predicted by the WCS were not significantly different \textit{($t(17) = 0.51, p = 0.62$)}. As predicted by the WCS, the sensitivity of the dyads whose members had similar (dissimilar) sensitivity levels was significantly higher (lower) than the relative sensitivity of dyad members (Figure \ref{fig:figure3}). We found a significant linear correlation between improvement of the dyads against the relative sensitivities of their members \textit{($R^2=0.62, F(1,17)=24.8, p=0.0002$)}, with slope (0.64 $\pm$ 0.13) and intercept ($0.66 \pm 0.09$) close to those predicted by the WCS model ($0.71$ for slope and intercept). 

\textcolor{black}{The results of our study align very well with those of Bahrami et al, \cite{bahrami2010}, who used explicit verbal communication -- and same WCS model applies equally well to both studies. However, group decision time was significantly faster in our study (N=850, mean=2856ms, std=2022ms) than in Bahrami et al, \cite{bahrami2010} (N=5 groups, mean=13860ms, std=3720ms; Dan Bang, personal communication).}


\subsection{Dyad members implicitly communicate and share their confidence by manipulating both the speed and the force of their movements}



The WCS model requires individual confidence levels to be shared (because they need to be integrated). This raises the question of how, in our study, participants (implicitly) communicate and negotiate through the haptic channel. Previous studies showed that dyadic sensorimotor tasks, such as tapping in synchrony or lifting objects together,  promote the emergence of Leader and Follower roles -- with the Leader determining (for example) the pace of joint tapping \cite{Pezzulo2018c}. Furthermore, Leaders often modify their action kinematics in communicative ways, to signal their roles and convey relevant task information to Followers \cite{Pezzulo2013,Sacheli2013,Candidi2015}. In keeping with this body of evidence, we asked whether participants exploit their kinematic and kinetic movement parameters (e.g., the speed and force of their movements) to become ``Leaders'' of the group decision and to implicitly communicate their leadership and confidence. 

We considered various kinematic and kinetic movement parameters as predictors of group choices. \textcolor{black}{First, we considered individuals' initial decision time for each trial as a predictor of leadership in the same trial}. The correlation between being the individual who moves first and being the Leader (i.e., determining the group choice) is 66.5\% overall (67.7\% and 63\% for dyads having similar and different sensitivities, respectively). 


As a further proxy to initiative and early commitment to the group decision, we designed a First Crossing (1C) predictor: the side (left or right) at which any of the participants' handles firstly exits a ``small zone'' centred on the start position \cite{roche2016}. We parametrised the ``small zone'' around the start position $X_0$ as $X_0 \pm X_{thresh}$ using different thresholds (see Table~\ref{table:gabor_1C}). We found that the side selected by the (first) participant who moves less than 10\% of the total distance to the target (i.e., threshold $X_{thresh} = 0.05$) already predicts 88.5\% of the group choices. 

%





Finally, we considered two kinetic parameters available through the haptic device -- peak force (i.e., the highest force applied by a subject on the interface) and mechanical work (i.e., $W_i = \frac{1}{N}\sum\limits_{k=1}^n F_i(X_{i,k}-X_{i,k-1})$, i=$\{$0,1$\}$, where $F_i$ is the force applied on interface $i$ and $X_{i,k}$ is the position of the interface $i$ at time step $k$.) -- and found that both are good predictors of group choices (71,7\% and 69\% accuracy, respectively). Leaders were more active than Followers during the interaction, with both significantly higher peak forces applied (Leader: 0.75N vs Follower: 0.43N; t(676)=9.71, p$<$0.0001) and significantly higher mechanical work provided (Leader: 0.30J vs Follower: -0.08J; t(676)=15.7, p$<$0.0001). Rather, Followers tended to apply negative mechanical work, effectively exerting some (small) resistance to the Leader's motion -- see \cite{melendez2011,reed2006} for similar results on dyadic co-manipulation.

\subsection{Leaders impose their pace to the group decision movements}


Previous studies of sensorimotor communication reported that Followers tend to align to Leaders' movements during joint tasks (see \cite{Pezzulo2018c} for a review). In keeping, we asked whether Leaders imposed their pace to the group decision movements (and Followers adapted to it). We considered the ratio between the mean velocity of Leaders (VeloL) and Followers (VeloF) during the first part of the dyad movement (i.e., before the $X_{Thresh}$ is crossed for the first time), which is arguable more important to reach consensus; and the mean velocity of the dyad (VeloD) during the second part of dyad movement (i.e., after the $X_{Thresh}$ is crossed the first time), which is necessary to complete the trial. We found VeloL/VeloD (1.0788) to be significantly smaller than VeloF/VeloD (1.1115), (N = 1866.0, p-value: 0.00348, t-value: -2.92344, d-value: -0.05613). The fact that the former ratio (VeloL/VeloD) is closer to 1 than the latter (VeloF/VeloD) ratio indicates that the Leader is more able to impose his pace on the group decision movements.

\subsection{Control analyses}


To rule out the possibility that group decisions were made without negotiation, by simply following ``who moves first (or pulls harder)'', we performed two control analyses, which compare individual and group decisions. We found group decision time (N=850, mean=2856ms, std=2022ms) to be significantly longer than decision time of group members decision time (N=4352, mean=881ms, std=788ms): t(850, 4352)=-23.84, p$<$0.0001. This result holds true both for dyads of similar (t(626,3328) = -23.42, p$<$0.0001) and different (t(224, 1024) = -10.15, p$<$0.0001) sensitivities. Furthermore, we found group initiation time, as indexed by the time the group reaches the 1C parameter (i.e., the handle firstly exit the starting zone $X_0$ as $X_0 \pm X_{thresh}$) to be significantly slower than initiation time of group members, but only for $T_{thresh}=$0.05 and for dyads having similar sensitivities  ($t(626,3328)=4.51, p<0.0001$). Rather, we found group initiation time to be significantly faster than initiation time of group members ($t(224, 1024)=-4.78, p<0.0001$) for dyads having different sensitivities. These two control analyses reassuringly suggest that group decisions require involve time-consuming negotiation; and the slower initiation time of groups whose members have similar sensitivities may be conducive of better choices.



\section*{Discussion}


Group decision making is an active area of research across behavioral sciences, psychology and neuroscience \cite{Toyokawa2014}, ecology \cite{Marshall2017} and collective (or swarm) robotics \cite{Rosenberg2016}; but its dynamics and optimality principles are still incompletely known.

We show that sensorimotor (haptic) communication can optimise group decisions. Haptically coupled dyads perform significantly better than the best individual of the dyad, when the two individual members have similar visual sensitivities; but the opposite is true when the dyad members have different sensitivity levels. This group advantage was shown in tasks using \emph{explicit}, verbal communication \cite{bahrami2010}. Here we demonstrate that \emph{implicit} (haptic) channels \textcolor{black}{can achieve the same results, at least in the joint decision investigated here}. 

\textcolor{black}{In general, verbal communication can be much richer than implicit communication. However, in the context of this task, the specific information to be conveyed concerns (one's belief about) the correct target. Both verbal and haptic channels can convey this information; but it is plausible that haptic information can convey it more precisely, i.e., with a better information/noise ratio. This speaks to the fact that explicit and implicit communication channels may have complementary benefits. Tasks requiring sophisticated debate and diplomacy and where response options are open-ended, such as legal proceedings or diplomatic negotiates, may be more difficult to address using implicit compared to explicit, verbal communication. On the other hand, simpler tasks having clear response options that can be mapped to spatial locations, implicit communication may afford a faster but equally accurate consensus compared to explicit communication.}

Indeed, in our study, consensus was reached (in most cases) in less than 3 seconds, whereas with verbal communication it required about 14 seconds \cite{bahrami2010}. Note that neither in our study nor in those using explicit communication there was any time pressure. Clearly, the comparison may seem unfair, as (compared to haptics) language is a much richer communication channel; and hence consensus and conventions may take time to arise. What is most interesting in this comparison is that implicit channels afford very fast group consensus -- which is relevant when it is necessary to trade off richness and speed of communication, such as during situated group decisions and team sports.

Our results can be explained within an optimal multisensory integration framework \cite{Ernst2002,Kording2006,Doya2007}, which combines multiple sources of evidence and weights them in proportion to their confidence levels. Importantly, the same WCS model that incorporates the above Bayesian assumptions explains group decisions that use both explicit \cite{bahrami2010} and implicit communication (this study), suggesting that the differences between the two may be less prominent than currently believed -- at least for the choices considered here. The computational model further indicates confidence sharing as a key ingredient to optimize group decisions. This raises the question of how exactly dyads communicate their confidence levels through the haptic channel. 


Our results show that co-actors share their confidence levels and optimise group decisions by synergistically manipulating the movement parameters that are available via the haptic interface, such as speed and force. We found initiative (as indexed by 1C) to be the most effective group choice predictor; indeed, the participant who takes the initiative and commits early to a decision often acquires leadership and determines the final choice. However, the fact that higher levels of activity and force afford accurate predictions of group choice indicates that both kinematic and kinetic parameters may be used in combination. Supposedly, an effective sensorimotor communication strategy consists in taking initiative and then applying some force to maintain -- and communicate -- commitment to the choice and to impose a pace to the decision. 

The success of such sensorimotor communication strategy may be due to the fact that both speed and force of movement reliably signal confidence in addition to advancing the decision. Indeed, given that response time and confidence are inversely correlated \cite{Pleskac2010,Vickers1982}, making an early commitment is a reliable signal that one is confident about the decision. Exerting force can reliably signal one's confidence, too. The parallel between force and confidence is made apparent by the recent finding that participants showing (sub-threshold) motor activation in their response effectors have significantly higher confidence (but not necessarily accuracy) in their choices \cite{Gajdos2019}. These findings suggest that haptic interfaces provide efficient channels, such as speed and force, for implicit \emph{confidence sharing}, which is key to optimal group decisions. 


Our results have deep theoretical and technological implications. From a theoretical perspective, our findings run against the hypothesis that explicit communication is necessary to achieve optimal decisions or to communicate confidence; suggesting that classical theories of communication should be expanded to consider more fully sensorimotor exchanges \cite{Donnarumma2017a,Donnarumma2017b,Donnarumma2017c,Metcalfe1996,Pezzulo2011,Pezzulo2011g,Pezzulo2013c,Sperber1995}. \textcolor{black}{From a technological perspective, our results can pave the way to the development of novel negotiation and decision support tools that exploit fast sensorimotor channels to facilitate and improve group decisions. While we focused on a visuo-haptic interface, similar results may be obtained using other (e.g., auditory) channels, to the extent that they afford rich sensorimotor communication. Our study reveals that crucial information for accurate group decisions -- one's own confidence about the decision -- can be conveyed by modulating speed and force of movement. It remains to be studied whether using different sensorimotor channels or interfaces offers the same or different ways to communicate confidence and other important information. Another} challenge for the future consists in expanding the scope of sensorimotor technologies, to afford more complex negotiation dynamics that may be required to reach consensus beyond simple tasks.






\begin{methods}

\subsection{Participants}

\textcolor{black}{Thirty-six participants (11 women) were recruited for this experiment amongst interns, master degree students, PhD students, post-docs and engineers of the Sorbonne Universit\'{e} and paired in dyads (8 M-M, 9 M-F, 1 F-F). Dyads were formed by pairing participants who were not friends or collaborators, to avoid possible influences of previous interactions on task performance and leadership.} Participants were free of any known psychiatric or neurological symptoms, non-corrected visual or auditory deficits and recent use of any substance that could impede concentration. They were all right handed. Their mean age was 26.3 (SD = 5.25). This research was reviewed and approved by the High Council for Research and Higher Education (HCERES) institutional ethics committee. The research was performed in accordance with the relevant guidelines and regulations. Informed consent was obtained from each participant. One dyad had to be excluded because one of the members systematically defaulted to her partner's choice in the second phase. The analysis was thus conducted on 34 participants.

\subsection{Procedure}

Dyad members are in the same testing room, seated side by side, and each has a computer screen. An opaque curtain is positioned between them in order to prevent them from seeing each other. Participants are instructed to refrain from trying to communicate orally with their partners for the duration of the experiment. Headphones playing pink noise are used to prevent the subjects from hearing each other or potential audio clues in the testing room. Visual feedback is provided to the participants through individual displays.

Each subject controls a custom, one degree-of-freedom haptic interface, which use two MAXON DC Motors (RE65-250W), connected to a 80mm handle for actuation and a magnetic encoder (CUI INC AMT11) \cite{roche-erts2018}. The full design of the haptic interface is open source, available on GitHub at: github.com/LudovicSaintBauzel/teleop-controller-bbb-xeno.git. 

Each experiment includes 8 blocks of 16 trials each. Subjects switch their positions after half the trials. Each trial proceeds as follows. First, the haptic interfaces are automatically centred and a warning message is displayed (1000ms). Second, a black central fixation cross is displayed on each subject's screen for a random duration (500-1000ms).  

The third phase is the \emph{individual decision phase}. Two visual stimuli (6 Gabor patches displayed in circle) are sequentially presented to both subjects, for 85ms. A 1000 ms pause (grey screen, black fixation cross) is observed between the two stimuli. In either the first or second stimulus, one of the 6 patches has a slightly higher contrast (oddball target). The task objective is to determine whether the oddball target is in the first or second stimulus. Note that he oddball targets can have 4 different level of contrast compared to the baseline. The oddball target timing (first or second wave), position (one of the six patches) and contrast (one of the four levels, $11.5\%$, $13.5\%$, $17\%$ and $25\%$; baseline is $10\%$) are randomized for each trial. The oddball timing and contrast levels were used as independent variables and the number of occurrences of each of their combinations was balanced over each block (each of the 8 combinations appear twice per block, for a total of 16 trials per block). After the presentation of the stimuli, both subjects must indicate their individual answer, by moving the handle of the haptic interface towards the left (to select the first stimulus) or right (to select the second stimulus). In this phase, the positions of the haptic interfaces are independent, and each subject answers individually. After both subjects have answered, both answers are displayed for each subject. If they agree, feedback about the correct answer is given with both a color code (green for a correct answer, red for an incorrect one) and a symbol (green check mark for a correct answer, red cross for an incorrect one); and the trial ends. If they disagree, only their individual choices is provided and participants enter in the \emph{group decision phase}.

%

The fourth phase is the \emph{group decision phase}. In this phase, haptic feedback is added to the interfaces: the teleoperation controller will constrain the motions of the interfaces so that there are identical at all time. In this configuration, the interfaces' positions are the same and the subjects have equal control over it. Furthermore, they can feel the force applied to the interfaces by each other. The subject must jointly move the interfaces in order to indicate their final choice (left for first stimulus, right for second). The interfaces must remain one second at stop in order to validate the common answer. During the group decision, participants can sense the force applied by their co-actors via the haptic interface. To avoid conflicts being resolved by brute force, participants are instructed to keep their force below the maximum.




Finally, after the group decision, feedback about the individual choices and the common decision are given to the subjects (CORRECT/WRONG). Feedback is color-coded: yellow for subject 1 on the left, blue for subject 1 on the right. Note that each feedback phase lasts a maximum of 10 seconds; but after 3 seconds, participants can skip by placing their fingers on the interface. At the end of the feedback, the graphical interface goes back to step 1, and the trials continue until the end of the experimental block.


%


%




\subsection{Psychometric functions}

Individual and dyadic psychometric functions are constructed by plotting the proportion of trials in which the oddball target was seen in the second wave of stimuli against the contrast difference at the oddball location (contrast in the second wave minus contrast in the first); see also \cite{bahrami2010}.

Examples of psychometric functions are shown in the main article. The dots correspond to the average proportion of 2$^{nd}$ stimuli chosen as answer, for each contrast difference ($\pm$ 1,5\%, $\pm$ 3.5\%, $\pm$ 7\%, $\pm$ 15\%). Lines are the fitted cumulative Gaussian functions for each individuals and dyads. The psychometric curves are fit to a cumulative Gaussian function whose parameters are bias (b) and variance ($\sigma ^2$). Estimation of these parameters is done through curve fitting regression (Python Scipy curve\_fit() function).

A participant with bias $b$ and variance $\sigma ^2$ would have a psychometric curve given by:

\begin{equation}
P(\Delta C) = H\left(\frac{\Delta C + b}{\sigma}\right),
\end{equation}

with $\Delta C$ the contrast difference between second and first stimuli, and H(z) the cumulative normal function.

The psychometric curve, P($\Delta$C), corresponds to the probability of reporting that the second stimulus had the higher contrast. Thus, a positive bias indicates an increased probability of saying that the second stimulus had higher contrast (and thus corresponds to a negative mean for the underlying Gaussian distribution).

Given the above definitions for P($\Delta$C), the variance is related to the maximum slope of the psychometric curve, denoted $s$, via :

\begin{equation}
s = \frac{1}{\sqrt{2\pi\sigma ^2}}.
\end{equation}

A large slope indicates small variance and thus highly sensitive performance. 


\subsection{Weighted Confidence Sharing (WCS) Model}
\label{sec:WCS}

We used the Weighted Confidence Sharing (WCS) model \cite{bahrami2010} to fit our behavioral data. The WCS assumes that participants share their confidence and make a Bayes-optimal decision based on the ratio of the individual $\delta C / \sigma$ values. This permits inferring the dyad psychometric function from the psychometric functions of the dyad members, as follows:

\begin{equation}
P_{dyad}^ { W C S }(\Delta C) = H\left(\frac{\Delta C + b_{dyad}^ { W C S }}{\sigma_{dyad}^ { W C S }}\right),
\end{equation}

with

\begin{equation}
b_{dyad}^ { W C S } = \frac{\sigma_2b_1 + \sigma1b_2}{\sigma1 + \sigma_2}
\end{equation}

and 

\begin{equation}
\sigma _ { d y a d } ^ { W C S } = \sqrt{2} \frac { \sigma _ { 1 } \sigma _ { 2 } } { \sigma _ { 1 } + \sigma _ { 2 } }.
\end{equation}

Consequently, the slope of the dyad's psychometric function can be calculated as: 

\begin{equation}
\label{eq:WCS_slope}
s_{dyad}^ { W C S } = \frac{s_1+s_2}{\sqrt{2}}.
\end{equation}

The WCS model predicts that the performance (sensitivity) of the dyad is superior to those of the best member if the sensitivities of the participants are similar. This can be appreciated by noting that if $s_{max}$ is the slope of the psychometric function of the best performing member of the dyad, and $s_{min}$ the slope of his/her partner's psychometric function, we have:

\begin{equation}
s_{dyad} = \frac{s_{min}+s_{max}}{\sqrt{2}} = \frac{1+\frac{s_{min}}{s_{max}}}{\sqrt{2}}s_{max}.
\end{equation}

If we compare the performances of the dyad and of the best performing member we have:

\begin{equation}
\label{eq:WCS_amelio}
\frac{s_{dyad}}{s_{max}} = \frac{1+\frac{s_{min}}{s_{max}}}{\sqrt{2}} = \frac{\sqrt{2}}{2} + \frac{\sqrt{2}}{2}\frac{s_{min}}{s_{max}}.
\end{equation}

The WCS model makes the theoretical prediction that if the ratio of sensitivities of the dyad's members is greater than 0.4 (i.e., participants have similar sensitivities), then the dyad will outperform each individual. The opposite happens if the ratio is lower than 0.4 (i.e., participants have different sensitivities). More formally, the following property holds: $s_{dyad} > s_{max}$ if $s_{min}/s_{max} > 1-\frac{\sqrt{2}}{2} \simeq 0.4$.


Our findings reported in the main article confirm the theoretical predictions of the WCS model and permit to rule out alternative models considered in \cite{bahrami2010}: the \emph{Coin Flip (CF)} model, which considers that conflicts are decided by chance; the \emph{Behaviour and Feedback (BF)} model, which considers that participants learn who is the most accurate group member and rely on his or her choice during group decisions; and the \emph{Direct Signal Sharing (DSS)} model, which considers that dyads communicate the mean and standard deviation of each member's sensory response. None of these alternative models would predict the pattern of responses that we observe in our data. 

Note that as reported in the main article, we found the slope of the linear regression fit of participants' performance to be slightly slower than what predicted by the WCS model (despite the difference does not reach significance). This finding can be explained with a small modification of the WCS model: by adding a small bias to the most skilled member of the dyad. Indeed, according to the WCS model, the dyad sensitivity improvement can be calculated as: $\frac{s_{dyad}}{s_{max}} = \frac{1}{\sqrt{2}}\left(1+\frac{s_{min}}{s_{max}}\right) = \frac{\sqrt{2}}{2} + \frac{\sqrt{2}}{2}\frac{s_{min}}{s_{max}}$ (Equation \ref{eq:WCS_amelio}). If the dyad slightly over-weights the decision of the most skilled member, the resulting sensitivity will be shifted towards $s_{max}$: 

\begin{equation}
\frac{s_{dyad}}{s_{max}} = \frac{\sqrt{2}}{2} + \frac{\sqrt{2}}{2}\frac{\alpha s_{min}}{\beta s_{max}},
\end{equation}

with $\beta > \alpha > 0$ the relative weights.

This model would lead to a similar intercept than the WCS model, with a lower slope, which would explain the pattern of results we obtain. 


\subsection{First Crossing (1C) parameter}


The 1C parameter is defined as the side on which the individual position of one of the two subjects exits the interval $[-X_{thresh}; X_{thresh}]$. The position data from the group decision phase are extracted and normalised so that the middle starting position corresponds to $X_{pos}=0$, and the left and right sides corresponds to $X_{pos}=-1$ and $X_{pos}=1$ respectively. The value of $X_{thresh}$ for the 1C calculations is then chosen as a percentage of $X_{pos}$.




\subsection{Mechanical work parameter}

The mechanical work of a subject is calculated as $W_i = \frac{1}{N}\sum\limits_{k=1}^n F_i(X_{i,k}-X_{i,k-1})$, i=$\{$0,1$\}$, where $F_i$ is the force applied on the interface $i$ and $X_{i,k}$ is the position of the interface $i$ at time step $k$.

\end{methods}





\begin{addendum}
 \item Funding: This work has received fundings from a grant to G.P. from the European Research Council (Grant Agreement No. 820213, ThinkAhead) and a grant to G.P from ONR (grant number N62909-19-1-2017). We thank Bahador Bahrami and Dan Bang for sharing data on their group decision tasks.
 \item[Competing Interests] The authors declare that they have no
competing financial interests.
\item[Correspondence] Correspondence and requests for materials
should be addressed to Giovanni Pezzulo~(email: giovanni.pezzulo@istc.cnr.it).
\item[Author contributions] GP, LR, LSB designed the research; LR conducted the research under the supervision of GP and LSB; GP, LR and LSB wrote the article.
\end{addendum}

\newpage


\begin{table}[h]
\center
\begin{tabular}{rccccccc}
\hline
$X_{thresh}$ :  & 0.05 & 0.08 & 0.10 & 0.15 & 0.20 & 0.25 & 0.30 \\
\hline
\% of correct predictions :  & 88.5 & 90.0 & 91.9 & 92.9 & 93.7 & 94.6 & 95.7 \\
\hline
\end{tabular}
\caption{Proportion of group choices correctly predicted by the 1C predictor.}
\label{table:gabor_1C}
\end{table}

\newpage


\newpage

\newpage

\end{document}